\documentstyle[11pt,fullname]{article}

\title{Incremental Parser Generation for Tree Adjoining
  Grammars\thanks{Thanks to Dania Egedi, Aravind Joshi, B. Srinivas
    and the student session reviewers.}}
\author{Anoop Sarkar \\
  University of Pennsylvania\\
  Department of Computer and Information Science\\
  Philadelphia PA 19104\\
  {\tt anoop@linc.cis.upenn.edu}} \date{2 February 1996}

\begin{document}

\catcode`\@=11\relax
\newwrite\@unused
\def\typeout#1{{\let\protect\string\immediate\write\@unused{#1}}}
\typeout{psfig/tex 1.2-dvips}


\def\figurepath{./}
\def\psfigurepath#1{\edef\figurepath{#1}}

%
%
\def\@nnil{\@nil}
\def\@empty{}
\def\@psdonoop#1\@@#2#3{}
\def\@psdo#1:=#2\do#3{\edef\@psdotmp{#2}\ifx\@psdotmp\@empty \else
    \expandafter\@psdoloop#2,\@nil,\@nil\@@#1{#3}\fi}
\def\@psdoloop#1,#2,#3\@@#4#5{\def#4{#1}\ifx #4\@nnil \else
       #5\def#4{#2}\ifx #4\@nnil \else#5\@ipsdoloop #3\@@#4{#5}\fi\fi}
\def\@ipsdoloop#1,#2\@@#3#4{\def#3{#1}\ifx #3\@nnil 
       \let\@nextwhile=\@psdonoop \else
      #4\relax\let\@nextwhile=\@ipsdoloop\fi\@nextwhile#2\@@#3{#4}}
\def\@tpsdo#1:=#2\do#3{\xdef\@psdotmp{#2}\ifx\@psdotmp\@empty \else
    \@tpsdoloop#2\@nil\@nil\@@#1{#3}\fi}
\def\@tpsdoloop#1#2\@@#3#4{\def#3{#1}\ifx #3\@nnil 
       \let\@nextwhile=\@psdonoop \else
      #4\relax\let\@nextwhile=\@tpsdoloop\fi\@nextwhile#2\@@#3{#4}}
\def\psdraft{
	\def\@psdraft{0}
}
\def\psfull{
	\def\@psdraft{100}
}
\psfull
\newif\if@prologfile
\newif\if@postlogfile
\newif\if@noisy
\def\pssilent{
	\@noisyfalse
}
\def\psnoisy{
	\@noisytrue
}
\psnoisy
\newif\if@bbllx
\newif\if@bblly
\newif\if@bburx
\newif\if@bbury
\newif\if@height
\newif\if@width
\newif\if@scale
\newif\if@rheight
\newif\if@rwidth
\newif\if@clip
\newif\if@verbose
\def\@p@@sclip#1{\@cliptrue}


\def\@p@@sfile#1{\def\@p@sfile{null}%
	        \openin1=#1
		\ifeof1\closein1%
		       \openin1=\figurepath#1
			\ifeof1\typeout{Error, File #1 not found}
			\else\closein1
			    \edef\@p@sfile{\figurepath#1}%
                        \fi%
		 \else\closein1%
		       \def\@p@sfile{#1}%
		 \fi}
\def\@p@@sfigure#1{\def\@p@sfile{null}%
	        \openin1=#1
		\ifeof1\closein1%
		       \openin1=\figurepath#1
			\ifeof1\typeout{Error, File #1 not found}
			\else\closein1
			    \def\@p@sfile{\figurepath#1}%
                        \fi%
		 \else\closein1%
		       \def\@p@sfile{#1}%
		 \fi}

\def\@p@@sbbllx#1{
		\@bbllxtrue
		\dimen100=#1
		\edef\@p@sbbllx{\number\dimen100}
}
\def\@p@@sbblly#1{
		\@bbllytrue
		\dimen100=#1
		\edef\@p@sbblly{\number\dimen100}
}
\def\@p@@sbburx#1{
		\@bburxtrue
		\dimen100=#1
		\edef\@p@sbburx{\number\dimen100}
}
\def\@p@@sbbury#1{
		\@bburytrue
		\dimen100=#1
		\edef\@p@sbbury{\number\dimen100}
}
\def\@p@@sscale#1{
		\@scaletrue
		\count255=#1
   		\edef\@p@sscale{\number\count255}
}
\def\@p@@sheight#1{
		\@heighttrue
		\dimen100=#1
   		\edef\@p@sheight{\number\dimen100}
}
\def\@p@@swidth#1{
		\@widthtrue
		\dimen100=#1
		\edef\@p@swidth{\number\dimen100}
}
\def\@p@@srheight#1{
		\@rheighttrue
		\dimen100=#1
		\edef\@p@srheight{\number\dimen100}
}
\def\@p@@srwidth#1{
		\@rwidthtrue
		\dimen100=#1
		\edef\@p@srwidth{\number\dimen100}
}
\def\@p@@ssilent#1{ 
		\@verbosefalse
}
\def\@p@@sprolog#1{\@prologfiletrue\def\@prologfileval{#1}}
\def\@p@@spostlog#1{\@postlogfiletrue\def\@postlogfileval{#1}}
\def\@cs@name#1{\csname #1\endcsname}
\def\@setparms#1=#2,{\@cs@name{@p@@s#1}{#2}}
%
%
\def\ps@init@parms{
		\@bbllxfalse \@bbllyfalse
		\@bburxfalse \@bburyfalse
		\@heightfalse \@widthfalse
		\@scalefalse
		\@rheightfalse \@rwidthfalse
		\def\@p@sbbllx{}\def\@p@sbblly{}
		\def\@p@sbburx{}\def\@p@sbbury{}
		\def\@p@sheight{}\def\@p@swidth{}
		\def\@p@sscale{}
		\def\@p@srheight{}\def\@p@srwidth{}
		\def\@p@sfile{}
		\def\@p@scost{10}
		\def\@sc{}
		\@prologfilefalse
		\@postlogfilefalse
		\@clipfalse
		\if@noisy
			\@verbosetrue
		\else
			\@verbosefalse
		\fi
}
%
%
\def\parse@ps@parms#1{
	 	\@psdo\@psfiga:=#1\do
		   {\expandafter\@setparms\@psfiga,}}
%
%
\newif\ifno@bb
\newif\ifnot@eof
\newread\ps@stream
\def\bb@missing{
	\if@verbose{
		\typeout{psfig: searching \@p@sfile \space  for bounding box}
	}\fi
	\openin\ps@stream=\@p@sfile
	\no@bbtrue
	\not@eoftrue
	\catcode`\%=12
	\loop
		\read\ps@stream to \line@in
		\global\toks200=\expandafter{\line@in}
		\ifeof\ps@stream \not@eoffalse \fi
		\@bbtest{\toks200}
		\if@bbmatch\not@eoffalse\expandafter\bb@cull\the\toks200\fi
	\ifnot@eof \repeat
	\catcode`\%=14
}	
\catcode`\%=12
\newif\if@bbmatch
\def\@bbtest#1{\expandafter\@a@\the#1
\long\def\@a@#1
\long\def\bb@cull#1 #2 #3 #4 #5 {
	\dimen100=#2 bp\edef\@p@sbbllx{\number\dimen100}
	\dimen100=#3 bp\edef\@p@sbblly{\number\dimen100}
	\dimen100=#4 bp\edef\@p@sbburx{\number\dimen100}
	\dimen100=#5 bp\edef\@p@sbbury{\number\dimen100}
	\no@bbfalse
}
\catcode`\%=14
\def\compute@bb{
		\no@bbfalse
		\if@bbllx \else \no@bbtrue \fi
		\if@bblly \else \no@bbtrue \fi
		\if@bburx \else \no@bbtrue \fi
		\if@bbury \else \no@bbtrue \fi
		\ifno@bb \bb@missing \fi
		\ifno@bb \typeout{FATAL ERROR: no bb supplied or found}
			\no-bb-error
		\fi
		\count203=\@p@sbburx
		\count204=\@p@sbbury
		\advance\count203 by -\@p@sbbllx
		\advance\count204 by -\@p@sbblly
		\edef\@bbw{\number\count203}
		\edef\@bbh{\number\count204}
}
%
%
\def\in@hundreds#1#2#3{\count240=#2 \count241=#3
		     \count100=\count240	
		     \divide\count100 by \count241
		     \count101=\count100
		     \multiply\count101 by \count241
		     \advance\count240 by -\count101
		     \multiply\count240 by 10
		     \count101=\count240	
		     \divide\count101 by \count241
		     \count102=\count101
		     \multiply\count102 by \count241
		     \advance\count240 by -\count102
		     \multiply\count240 by 10
		     \count102=\count240	
		     \divide\count102 by \count241
		     \count200=#1\count205=0
		     \count201=\count200
			\multiply\count201 by \count100
		 	\advance\count205 by \count201
		     \count201=\count200
			\divide\count201 by 10
			\multiply\count201 by \count101
			\advance\count205 by \count201
		     \count201=\count200
			\divide\count201 by 100
			\multiply\count201 by \count102
			\advance\count205 by \count201
		     \edef\@result{\number\count205}
}
\def\compute@wfromh{
		\in@hundreds{\@p@sheight}{\@bbw}{\@bbh}
		\edef\@p@swidth{\@result}
}
\def\compute@hfromw{
		\in@hundreds{\@p@swidth}{\@bbh}{\@bbw}
		\edef\@p@sheight{\@result}
}
\def\compute@wfroms{
		\in@hundreds{\@p@sscale}{\@bbw}{100}
		\edef\@p@swidth{\@result}
}
\def\compute@hfroms{
		\in@hundreds{\@p@sscale}{\@bbh}{100}
		\edef\@p@sheight{\@result}
}
\def\compute@handw{
		\if@scale
			\compute@wfroms
			\compute@hfroms
		\else
			\if@height 
				\if@width
				\else
					\compute@wfromh
				\fi	
			\else 
				\if@width
					\compute@hfromw
				\else
					\edef\@p@sheight{\@bbh}
					\edef\@p@swidth{\@bbw}
				\fi
			\fi
		\fi
}
\def\compute@resv{
		\if@rheight \else \edef\@p@srheight{\@p@sheight} \fi
		\if@rwidth \else \edef\@p@srwidth{\@p@swidth} \fi
}
%
\def\compute@sizes{
	\compute@bb
	\compute@handw
	\compute@resv
}
%
%
\def\psfig#1{\vbox {
	%
	\ps@init@parms
	\parse@ps@parms{#1}
	\compute@sizes
	\ifnum\@p@scost<\@psdraft{
		\if@verbose{
			\typeout{psfig: including \@p@sfile \space }
		}\fi
		\special{ps::[begin] 	\@p@swidth \space \@p@sheight \space
				\@p@sbbllx \space \@p@sbblly \space
				\@p@sbburx \space \@p@sbbury \space
				startTexFig \space }
		\if@clip{
			\if@verbose{
				\typeout{(clip)}
			}\fi
			\special{ps:: doclip \space }
		}\fi
		\if@prologfile
		    \special{ps: plotfile \@prologfileval \space } \fi
		\special{ps: plotfile \@p@sfile \space }
		\if@postlogfile
		    \special{ps: plotfile \@postlogfileval \space } \fi
		\special{ps::[end] endTexFig \space }
		\vbox to \@p@srheight true sp{
			\hbox to \@p@srwidth true sp{
				\hss
			}
		\vss
		}
	}\else{
		\vbox to \@p@srheight true sp{
		\vss
			\hbox to \@p@srwidth true sp{
				\hss
				\if@verbose{
					\@p@sfile
				}\fi
				\hss
			}
		\vss
		}
	}\fi
}}
\def\psglobal{\typeout{psfig: PSGLOBAL is OBSOLETE; use psprint -m instead}}
\catcode`\@=12\relax

\maketitle

\begin{abstract}
  This paper describes the incremental generation of parse
  tables for the LR-type parsing of Tree Adjoining Languages (TALs).
  The algorithm presented handles modifications to the input grammar
  by updating the parser generated so far.  In this paper, a lazy
  generation of LR-type parsers for TALs is defined in which parse
  tables are created by need while parsing. We then describe an
  incremental parser generator for TALs which responds to modification
  of the input grammar by updating parse tables built so far.
\end{abstract}

\section{Introduction}

Tree Adjoining Grammars (TAGs) are tree rewriting systems which
combine trees with the single operation of {\it adjunction} (see
Figure~\ref{fig:tagdef}).  The construction of deterministic bottom-up
left to right parsing of Tree Adjoining Languages (TALs)%
\footnote{Familiarity with Tree Adjoining Grammars (TAGs) and their
  parsing techniques is assumed throughout the paper. For an
  introduction to TAGs, see \cite{Joshi87}. We shall assume that our
  definition of TAG does not have the {\it substitution} operation.
  Refer to \cite{Schabes91} for a background on the parsing of TAGs.}%
\cite{Schabes90} is an extension of the LR parsing
strategy for context free languages \cite{Aho86}. Parser generation
involves precompiling as much top-down information as possible into a
parse table which is used by the LR parsing algorithm.  This paper
gives an algorithm for the incremental generation of parse tables for
the LR-type parsing of TAGs.

\begin{figure}[htbp]
  \begin{center}
    \leavevmode
    \psfig{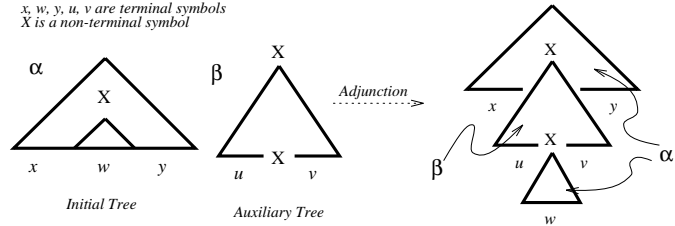}
    \caption{The Adjunction Operation}
    \label{fig:tagdef}
  \end{center}
\end{figure}

Parser generation provides a fast solution to the parsing of input
sentences as certain information about the grammar is precompiled and
available while parsing. However, if the grammar used to generate the
parser is either dynamic or needs frequent modification then the time
needed to parse the input is determined by both the parser and the
parser generator.

The main application area for TAGs has been the description of natural
languages. In such an area grammars are very rarely static, and
modifications to the original grammar are commonplace. In such an
interactive environment, conventional LR-type parsing suffers from the
following disadvantages:
\begin{itemize}
\item
Some parts of the grammar might never be used in the parsing of sentences
actually given to the parser. The time taken by the parser generator over
such parts is wasted.
\item
Usually, only a small part of the grammar is modified. So a parser generator
should also correspondingly make a small change to the parser rather than
generate a new one from scratch.
\end{itemize}

The algorithm described here allows the incremental incorporation of
modifications to the grammar in a LR-type parser for TALs. This paper extends
the work done on the incremental modification of LR(0) parser generators for
CFGs in \cite{Heering90,Heering89}. We define a lazy and incremental parser
generator having the following characteristics:
\begin{itemize}
\item The parse tables are generated in a lazy fashion from the
  grammar, i.e. generation occurs while parsing the input. Information
  previously precompiled is now generated depending on the input.
\item The parser generator is incremental. Changes in the grammar
  trigger a corresponding change in the already generated parser.
  Parts of the parser not affected by the modifications in the grammar
  are reused.
\item Once the needed parts of the parser have been generated, the
  parsing process is as efficient as a conventionally generated one.
\end{itemize}

Incremental generation of parsers gives us the following benefits:
\begin{itemize}
\item The LR-type parsing of lexicalized TAGs \cite{Schabes91}.  With
  the use of the lazy and incremental parser generation, lexicalized
  descriptions of TAGs can be parsed using LR-type parsing techniques.
  Parse tables can be generated without exhaustively considering all
  lexical items that anchor each tree.
\item Modular composition of parsers, where various modules of TAG
  descriptions are integrated with recompilation of only the necessary
  parts of the parse table of the combined parser.
\end{itemize}

\section{LR Parser Generation}

\cite{Schabes90} describe the construction of an LR
parsing algorithm for TAGs. Parser generation here is taken to be the
construction of LR(0) tables (i.e. without any lookahead) for a particular TAG%
\footnote{ The algorithm described here can be extended to a
  parser with SLR(1) tables~\cite{Schabes90}.}%
. The moves made by the parser can be most succinctly explained by
looking at an automaton which is weakly equivalent to TAGs called
Bottom-Up Embedded Pushdown Automata (BEPDA)~\cite{Schabes90}%
\footnote{ Note that the LR(0) tables considered here are
  deterministic and hence correspond to a subset of the TALs.
  Techniques developed in~\cite{Tomita86} can be used to resolve
  nondeterminism in the parser.}%
. The storage of a BEPDA is a sequence of stacks (or pushdown stores)
where stacks can be introduced above and below the top stack in the
automaton. Recognition of adjunction can be informally seen to be
equivalent to the {\bf unwrap} move shown in
Figure~\ref{fig:bepda-unwrap}.

\begin{figure}[htbp]
  \begin{center}
    \leavevmode
    \psfig{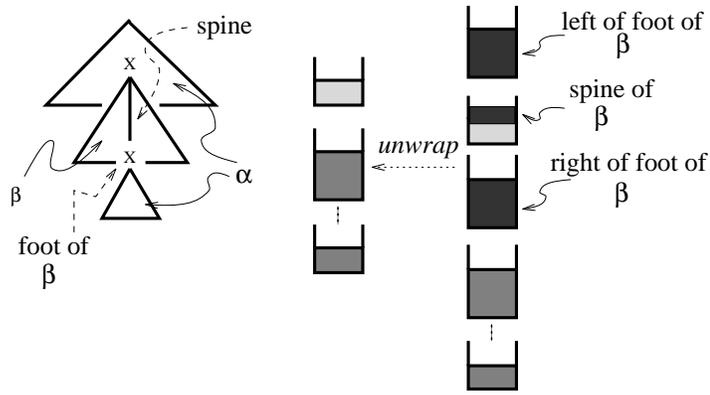}
    \caption{Recognition of adjunction in a BEPDA.}
    \label{fig:bepda-unwrap}
  \end{center}
\end{figure}

The LR parser uses a parsing table and a sequence of stacks (see
Figure~\ref{fig:bepda-unwrap}) to parse the input. The parsing table
encodes the actions taken by the parser as follows (with the help of
two $GOTO$ functions):
\begin{itemize}
\item {\bf Shift} to a new state which is pushed onto a new stack
  which appears on top of the current sequence of stacks.
\item {\bf Resume Right} where the parser has reached right and below
  a node on which an auxiliary tree has been adjoined.
  Figure~\ref{fig:resume-right} gives the two cases where the string
  beneath the foot node of an auxiliary tree has been recognized (in
  some other tree) and where the $GOTO_{foot}$ function encodes the
  proper state such that the right part of an auxiliary tree can be
  recognized.
\item {\bf Reduce Root} which causes the parser to execute an unwrap
  move to recognize adjunction (see Figure~\ref{fig:bepda-unwrap}).
  The proper state for the parser after adjunction is given by the
  $GOTO_{right}$ function.
\item {\bf Accept} and {\bf Error} functions as in conventional LR
  parsing.
\end{itemize}

\begin{figure}[htbp]
  \begin{center}
    \leavevmode
    \psfig{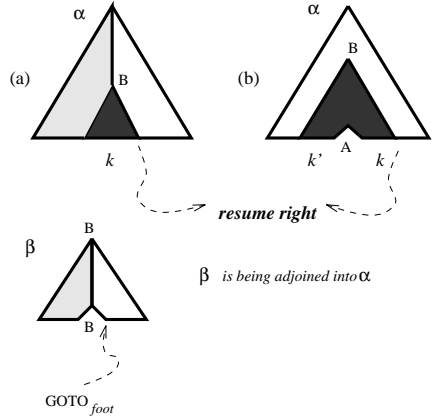}
    \caption{The {\bf resume right} action in the parser.}
    \label{fig:resume-right}
  \end{center}
\end{figure}

Figure~\ref{fig:tree-traversal} shows how the concept of dotted rules
for CFGs is extended to trees. There are four positions for a dot
associated with a symbol: left above, left below, right below and
right above. A dotted tree has one such dotted symbol. The tree
traversal in Figure~\ref{fig:tree-traversal} scans the frontier of the
tree from left to right while trying to recognize possible adjunctions
between the above and below positions of the dot. If an adjunction has
been performed on a node then it is marked with a star (e.g. $B*$).

\begin{figure}[htbp]
  \begin{center}
    \leavevmode
    \psfig{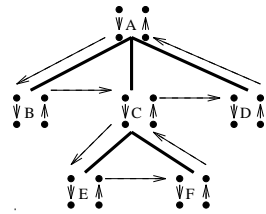}
    \caption{Left to right dotted tree traversal.}
    \label{fig:tree-traversal}
  \end{center}
\end{figure}

Construction of a LR(0) parsing table is an extension of the technique
used for CFGs.  The parse table is built as a finite state automaton
(FSA) with each state defined to be a set of dotted trees.  The
closure operations on states in the parse table are defined in
Figure~\ref{fig:closure-ops}. All the states in the parse table must
be closed under these operations. Figure~\ref{fig:table-2} is a
partial FSA constructed for the grammar in Figure~\ref{fig:grammar-1}.

\begin{figure*}[htbp]
  \begin{center}
    \leavevmode
    \psfig{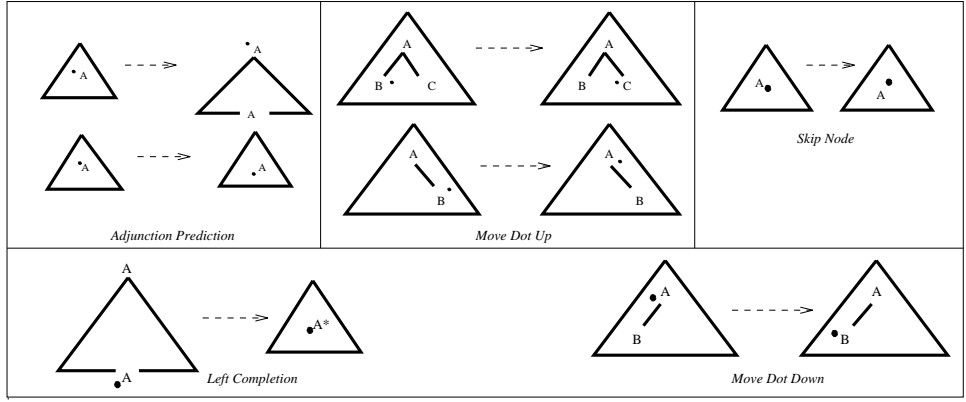}
    \caption{Closure Operations.}
    \label{fig:closure-ops}
  \end{center}
\end{figure*}

The FSA is built as follows: in state $0$ put all the initial trees
with the dot left and above the root. The state is then closed.  New
states are built by the transitions defined in
Figure~\ref{fig:transitions}. Entries in the parse table are
determined as follows:
\begin{itemize}
\item a {\bf shift} for each transition in the FSA.
\item {\bf resume right} iff there is a node $B*$ with the dot right
  and below it.
\item {\bf reduce root} iff there is a rootnode in an auxiliary tree
  with the dot right and above it.
\item {\bf accept} and {\bf error} with the usual interpretation.
\end{itemize}
The items created in each state before closure applies, i.e. the right
hand sides in Figure~\ref{fig:transitions} are called the {\bf
  kernels} of each state in the FSA. The initial trees with the dot
left and above the root form the kernel for state $0$. A state which
has not been closed is said to be in kernel form.

\begin{figure}[htbp]
  \begin{center}
    \leavevmode
    \psfig{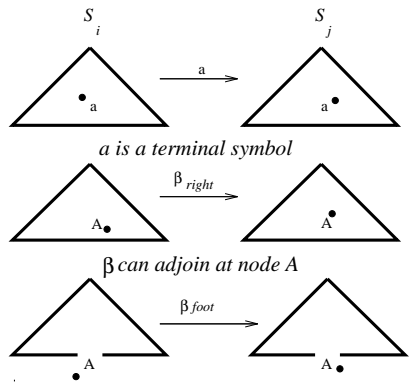}
    \caption{Transitions in the finite state automaton.}
    \label{fig:transitions}
  \end{center}
\end{figure}

\section{Lazy Parser Generation}

The algorithm described so far assumes that the parse table is
precompiled before the parser is used. Lazy parser generation spreads
the generation of the parse table over the parsing of several
sentences to obtain a faster response time in the parser generation
stage. It generates only those parts of the parser that are needed to
parse the sentences given to it. Lazy parser generation is useful in
cases where typical input sentences are parsed with a small part of
the total grammar.

We define lazy parser generation mainly as a step towards incremental
parser generation. The approach is an extension of the algorithm for
CFGs given in \cite{Heering90,Heering89}. To modify the LR parsing
strategy given earlier we move the closure and computation of
transitions (Figure~\ref{fig:closure-ops} and
Figure~\ref{fig:transitions}) from the table generation stage to the
LR parser.  The lazy technique expands a kernel state only when the
parser, looking at the current input, indicates that the state needs
expansion. For example, the TAG in Figure~\ref{fig:grammar-1} ($na$
rules out adjunction) produces the FSA in Figure~\ref{fig:table-1}%
\footnote{ As a convention in our FSAs we mark unexpanded kernel
  states with a boldfaced outline and a double-lined outline as the
  acceptance states.}%
. Computation of closure and transitions in the state occurs while
parsing as seen in Figure~\ref{fig:table-2} which is the result of the
LR parser expanding the FSA in Figure~\ref{fig:table-1} while parsing
the string $aec$. 

\begin{figure}[htbp]
  \begin{center}
    \leavevmode
    \psfig{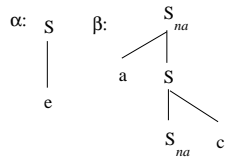}
    \caption{TAG $G$ where $L(G) = \{ a^n e c^n \}$}
    \label{fig:grammar-1}
  \end{center}
\end{figure}

\begin{figure}[htbp]
  \begin{center}
    \leavevmode
    \psfig{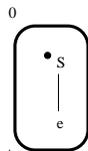}
    \caption{The FSA after parse table generation.}
    \label{fig:table-1}
  \end{center}
\end{figure}

\begin{figure*}[htbp]
  \begin{center}
    \leavevmode
    \psfig{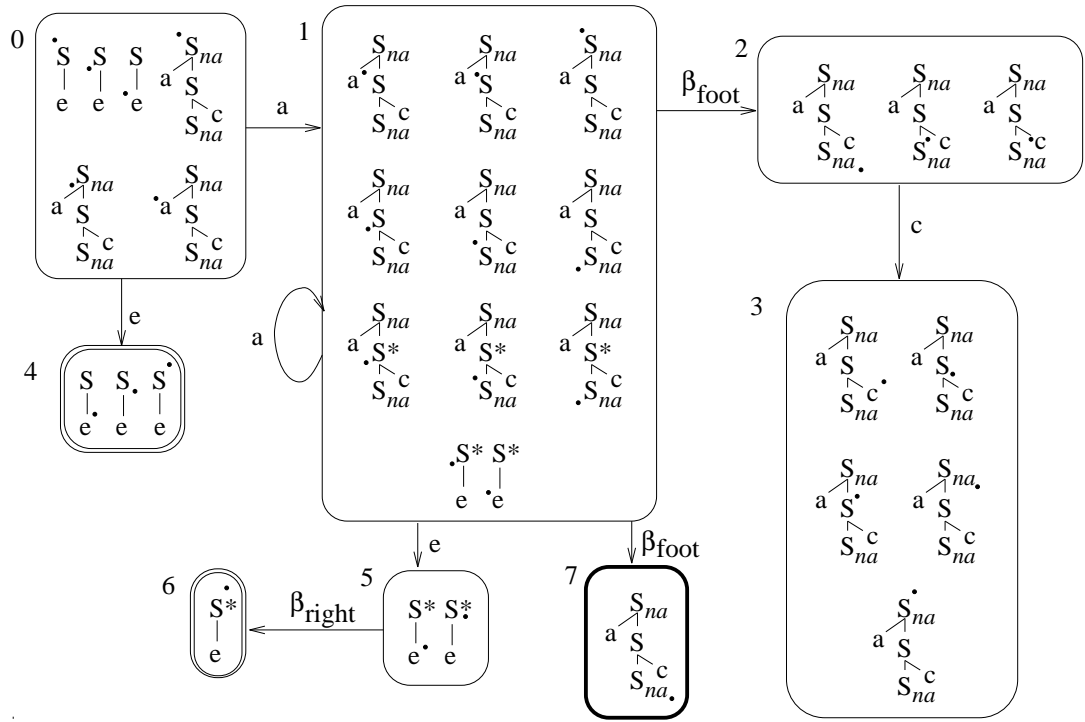}
    \caption{The FSA after parsing the string $aec$.}
    \label{fig:table-2}
  \end{center}
\end{figure*}

\begin{figure}[htbp]
  \begin{center}
    \leavevmode
    \psfig{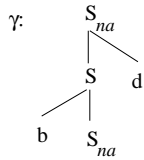}
    \caption{New tree added to $G$ with $L(G) = \{ a^n b^m e c^n d^m \}$}
    \label{fig:grammar-2}
  \end{center}
\end{figure}

The only extra statement in the modified parse function is a check on
the type of the state and possible expansion of kernel states takes
place while parsing a sentence.  Memory use in the lazy technique is
greater as the FSA is needed during parsing as well.

\section{Incremental Parser Generation}

The lazy parser generator described reacts to modifications to the
grammar by throwing away all parts of the parser that it has generated
and creates a FSA containing only the start state. In this section we
describe an incremental parser generator which retains as much of the
original FSA as it can. It throws away only that information from the
FSA of the old grammar which is incorrect with respect to the updated
grammar.

The incremental behaviour is obtained by selecting the states in the
parse table affected by the change in the grammar and returning them
to their kernel form (i.e. remove items added by the closure
operations). The parse table FSA will now become a disconnected graph.
The lazy parser will expand the states using the new grammar. All
states in the disconnected graph are kept as the lazy parser will
reconnect with those states (when the transitions in
Figure~\ref{fig:transitions} are computed) that are unaffected by the
change in the grammar. Consider the addition of a tree to the grammar%
\footnote{ Deletion of a tree will be similar. }%
.
\begin{itemize}
\item for an initial tree $\alpha$ return state 0 to kernel form
  adding $\alpha$ with the dot left and above the root node. Also
  return all states where a possible {\em Left Completion} on $\alpha$
  can occur to their kernel form.
\item for an auxiliary tree $\beta$ return all states where a possible
  {\em Adjunction Prediction} on $\beta$ can occur and all states with
  a $\beta_{right}$ transition to their kernel form.
\end{itemize}

For example, the addition of the tree in Figure~\ref{fig:grammar-2}
causes the FSA to fragment into the disconnected graph in
Figure~\ref{fig:table-3}. It is crucial that the disconnected states
are kept around as can be seen from the re-expansion of a single state
in Figure~\ref{fig:table-4}. All states compatible with the modified
grammar are eventually reused.

\begin{figure*}[htbp]
  \begin{center}
    \leavevmode
    \psfig{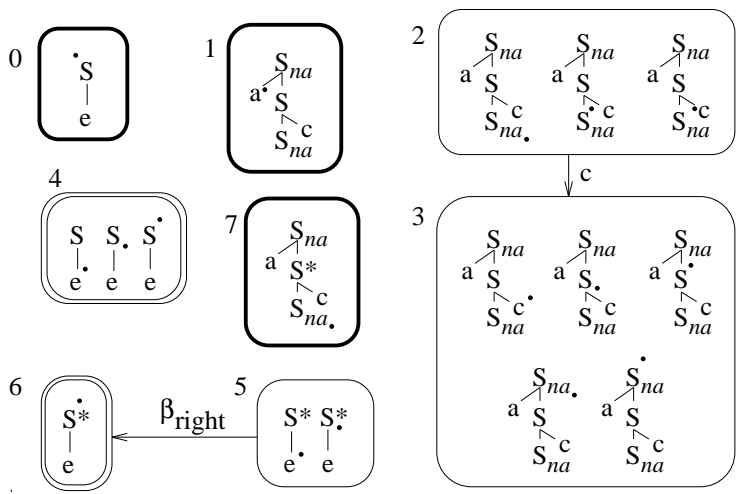}
    \caption{The parse table after the addition of $\gamma$.}
    \label{fig:table-3}
  \end{center}
\end{figure*}

\begin{figure*}[htbp]
  \begin{center}
    \leavevmode
    \psfig{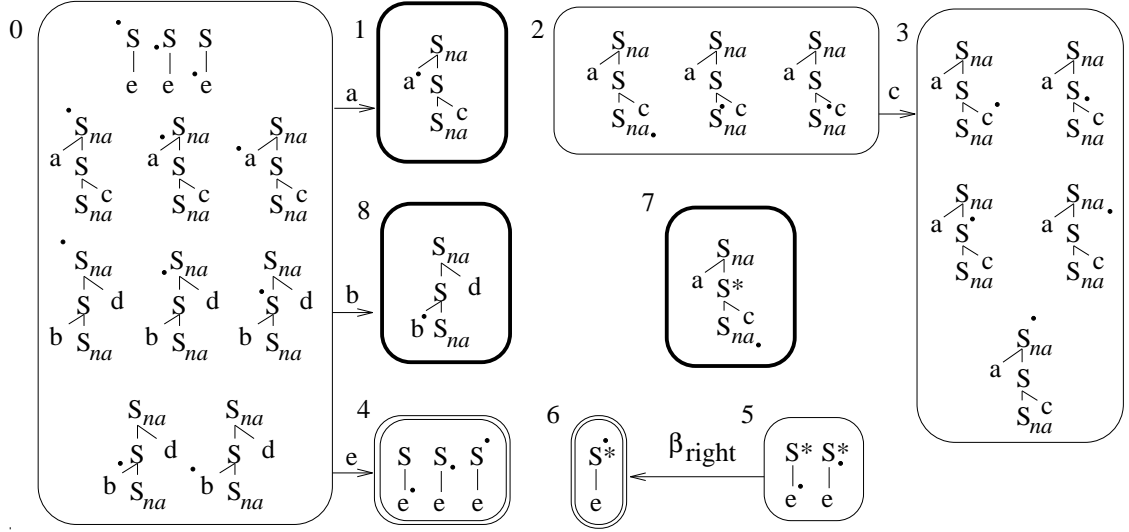}
    \caption{The parse table after expansion of state $0$ with the
      modified grammar.}
    \label{fig:table-4}
  \end{center}
\end{figure*}

The approach presented above causes certain states to become
unreachable from the start state. Frequent modifications of a grammar
can cause many unreachable states. A {\it garbage collection\/} scheme
defined in \cite{Heering90} can be used here which avoids
overregeneration by retaining unreachable states.

\section{Conclusion}

What we have described above is work in progress in implementing a
LR-type parser for a wide-coverage lexicalized grammar of English in
the TAG framework~\cite{xtag}. The algorithm for incremental parse
table generation for TAGs given here extends a similar result for
CFGs. The parse table generator was built on a lazy parser generator
which generates the parser only when the input string uses parts of
the parse table not previously generated. The technique for
incremental parser generation allows the addition and deletion of
elementary trees from a TAG without recompilation of the parse table
for the updated grammar. This allows us to combine the speed-up
obtained by precompiling top-down dependencies such as the prediction
of adjunction with the flexibility in lexical description usually
given by Earley-style parsers.

\bibliographystyle{fullname}

\end{document}